\documentclass[AMA,STIX1COL]{WileyNJD-v2}

\articletype{Article Type}%

\received{26 April 2016}
\revised{6 June 2016}
\accepted{6 June 2016}
\usepackage{booktabs,makecell,multirow}
\begin{document}

\title{Fixed-time Integral Sliding Mode Control for Admittance Control of a Robot Manipulator}

\author[1]{Yuzhu Sun}

\author[2]{Mien Van*}

\author[3]{Stephen McIlvanna}

\author[4]{Seán McLoone}

\author[5]{Dariusz Ceglarek}

\authormark{AUTHOR ONE \textsc{et al}}

\address[1]{\orgdiv{School of electronics, electrical and computer science}, \orgname{Queen's University of Belfast}, \orgaddress{\state{Belfast}, \country{UK}}}

\address[2]{\orgdiv{School of electronics, electrical and computer science}, \orgname{Queen's University of Belfast}, \orgaddress{\state{Belfast}, \country{UK}}}

\address[3]{\orgdiv{School of electronics, electrical and computer science}, \orgname{Queen's University of Belfast}, \orgaddress{\state{Belfast}, \country{UK}}}

\address[4]{\orgdiv{School of electronics, electrical and computer science}, \orgname{Queen's University of Belfast}, \orgaddress{\state{Belfast}, \country{UK}}}

\address[5]{\orgdiv{Warwick Manufacturing Group}, \orgname{University of Warwick}, \orgaddress{\state{Coventry}, \country{UK}}}

\corres{*Mien Van, School of electronics, electrical and computer science, Queen's University of Belfast, Belfast BT7 1NN. \email{M.Van@qub.ac.uk}}

\abstract[Summary]{This paper proposes a novel fixed-time integral sliding mode controller for admittance control to enhance physical human-robot collaboration. The proposed method combines the benefits of compliance to external forces of admittance control and high robustness to uncertainties of integral sliding mode control (ISMC), such that the system can collaborate with a human partner in an uncertain environment effectively. Firstly, a fixed-time sliding surface is applied in the ISMC to make the tracking error of the system converge within a fixed-time regardless of the initial condition. Then, a fixed-time backstepping controller (BSP) is integrated into the ISMC as the nominal controller to realize global fixed-time convergence. Furthermore, to overcome the singularity problem, a non-singular fixed-time sliding surface is designed and integrated into the controller, which is useful for practical application. Finally, the proposed controller is validated for a two-link robot manipulator with uncertainties and external human forces. The results show that the proposed controller is superior in the sense of both tracking error and convergence time, and at the same time, can comply with human motion in a shared workspace.}


\jnlcitation{\cname{%
\author{Sun Y.}, 
\author{Van M.}, 
\author{McIlvanna S.}, 
\author{McLoone S.}, and 
\author{Ceglarek D.}} (\cyear{2022}), 
\ctitle{Fixed-time Integral Sliding Mode Control for Admittance Control of a Robot Manipulator}, 
\cjournal{Manuscript submitted for publication.}.}

\maketitle


\section{Introduction}\label{sec1}

Recent developments in the biomedical, social and industrial robotics have led to an increasing interest in the field of physical human-robot collaboration (pHRC).\cite{9531394,he2020admittance} 
Robots that can physically collaborate with human partners enable the cognitive strengths of human and the high-precision and repeatability of robots to be combined. 
Unlike fully automated industrial production lines, to effectively work with the human partner in a shared workspace involving activities such as object handovers or co-carrying, it is imperative that the cooperative robots are capable of simultaneously complying with human forces at the contact point rather than rejecting them as external disturbances.\cite{tee2010adaptive} 
Therefore, impedance/admittance control based compliance control strategies have been attracting significant interest for this application. 
These include the adaptive admittance control,\cite{okunev2012human} adaptive impedance control,\cite{huang2004adaptive,tsumugiwa2002variable} and robust impedance control.\cite{chan1991robust,liu1991robust}

\begin{figure}[htb]
	\centering
	\includegraphics[width=6.5in]{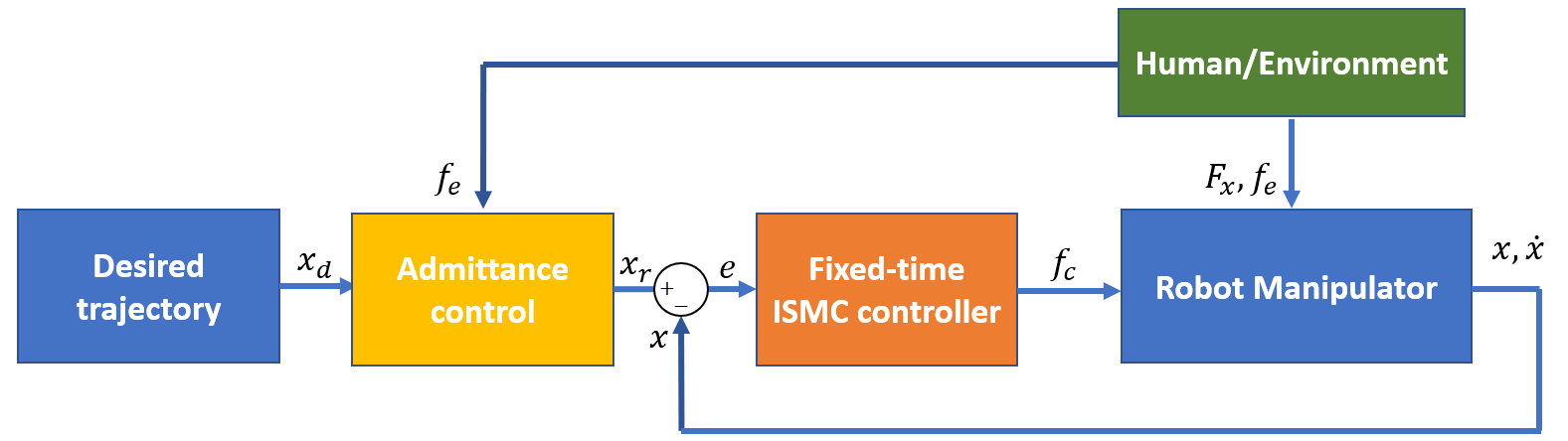}
	\caption{Structure of the proposed FTISMC Admittance Control Scheme.}
	\label{fig_1}
\end{figure}

Consider an industrial robot that is required to work with a human partner in an uncertain environment. The controller of the robot should be capable of tracking a certain trajectory with external uncertainties and perturbations, and at the same time, be compliant to human forces without the loss of stability. 
A considerable literature exists on techniques for improving the tracking performance of robot manipulators in an uncertain environment. 
For many years, model-free control strategies such as PID have been extensively employed in the industry.\cite{zuo2015non} Sliding mode control (SMC) based methods have also emerged and been widely studied in the field of robotics to provide robustness against the effects of parametric uncertainties and external disturbances.\cite{van2019adaptive} The SMC scheme involves (i) the reaching phase, in which the system converges from the initial state to the sliding surface, and (ii) the sliding phase, in which the system reaches the sliding surface and tries to stay on the equilibrium point. 
However, the effectiveness of SMC is limited by the reaching phase which is sensitive to disturbances. Therefore, to overcome the limitation of SMC, integral sliding mode control (ISMC) has been developed. ISMC eliminates the reaching phase by finding a suitable initial position to reject disturbances from the beginning, which is helpful for practical applications.\cite{van2019adaptive}

Furthermore, to obtain the property of fast tracking-error convergence, a further extension of ISMC, called fixed-time integral sliding mode control (FTISMC),\cite{huang2020robust,RN446,Wang_2018} which incorporates the benefits of ISMC and fixed-time sliding mode control (FxTSMC)\cite{shifix} has been proposed. Compared with traditional finite-time control strategies,\cite{yang2018adaptive,zuo2016distributed} the fixed-time controller does not depend on the initial condition of the system with the result that the convergence time can be determined a priori based on the design parameters.\cite{zuo2016distributed} 
In the work of Huang and Wang,\cite{huang2020robust} a FTISMC is proposed that stabilizes a highly complex nonlinear system within a fixed time and effectively restrains the irregular and nonlinear vibrations of the system. 
In the work of Li et al.\cite{RN446} a new integral high order sliding mode controller is proposed for high-order nonlinear systems with fixed-time convergence. 
In addition, the singularity problem always comes up in the sense of fixed-time stability when calculating the derivative of sliding variables. 
To address this issue a number of approaches have been developed to avoid or eliminate the singularity. \cite{feng2002non,yang2011nonsingular,zuo2015non,zuo2015nonsingular,van2021robust} 
In the work of Feng et al.\cite{feng2002non} a global non-singular terminal sliding mode controller is presented for second-order systems to overcome the singularity problem. 
In the work of Zuo,\cite{zuo2015nonsingular} a new sliding surface is proposed to circumvent the singularity problem. 
In the work of Van and Ceglarek,\cite{van2021robust} a new fault-tolerant control scheme based on a nonsingular fixed-time sliding mode controller is proposed for robot manipulators, which guarantees a global fixed-time convergence. 

So far there is little literature that combines admittance control and FTISMC for physical human-robot collaboration. Motivated by the above discussion this paper proposes a new ISMC control scheme for admittance control of a robot manipulator, as depicted in Fig. \ref{fig_1}. 
The stability and tracking error of the closed-loop system are converged within a fixed time. Furthermore, for practical application, a nonsingular fixed-time integral sliding mode controller is designed. The proposed method is then simulated on a two-link robot manipulator and compared with other state-of-the-art control strategies. The simulation results show that the proposed controller is superior in the sense of both tracking error and convergence time. The contributions and innovations of the proposed approach can be highlighted in a comparison with other approaches as follows:
\begin{enumerate}[1.]
	\item  Compared to conventional admittance control,\cite{he2020admittance} the proposed approach integrates an admittance with a FTISMC to guarantee a fixed-time convergence, while compliant with external forces. Therefore, the proposed controller provides faster transient response, lower tracking errors and better disturbance rejection capacity. 
	\item Compared to the existing FTISMC,\cite{huang2020robust,RN446,Wang_2018} which uses the tracking errors and the velocity of the tracking errors to reconstruct the integral sliding surface, the proposed controller develops a new nonsingular fixed-time sliding surface and uses the dynamics model of the system, which is reconstructed based on the nonsingular fixed-time sliding surface, to reconstruct a new integral sliding surface. Therefore, the proposed approach is able to avail of the full advantages of both ISMC and FxTSMC.
\end{enumerate}

The paper is organized as follows. The mathematical model describing the motion of a robot manipulator and the problem formulation are presented in Section \ref{sec2}.  The proposed fixed-time ISMC scheme for admittance control is developed in Section \ref{sec3}. Simulation results demonstrating the effectiveness of the proposed controller are presented in Section \ref{sec4}. Finally, section \ref{sec5} discusses the conclusions and directions for future works.

\section{Problem formulation and preliminaries}\label{sec2}

\subsection{Problem Formulation}
\begin{figure}
	\centering
	\includegraphics[width=3.9in]{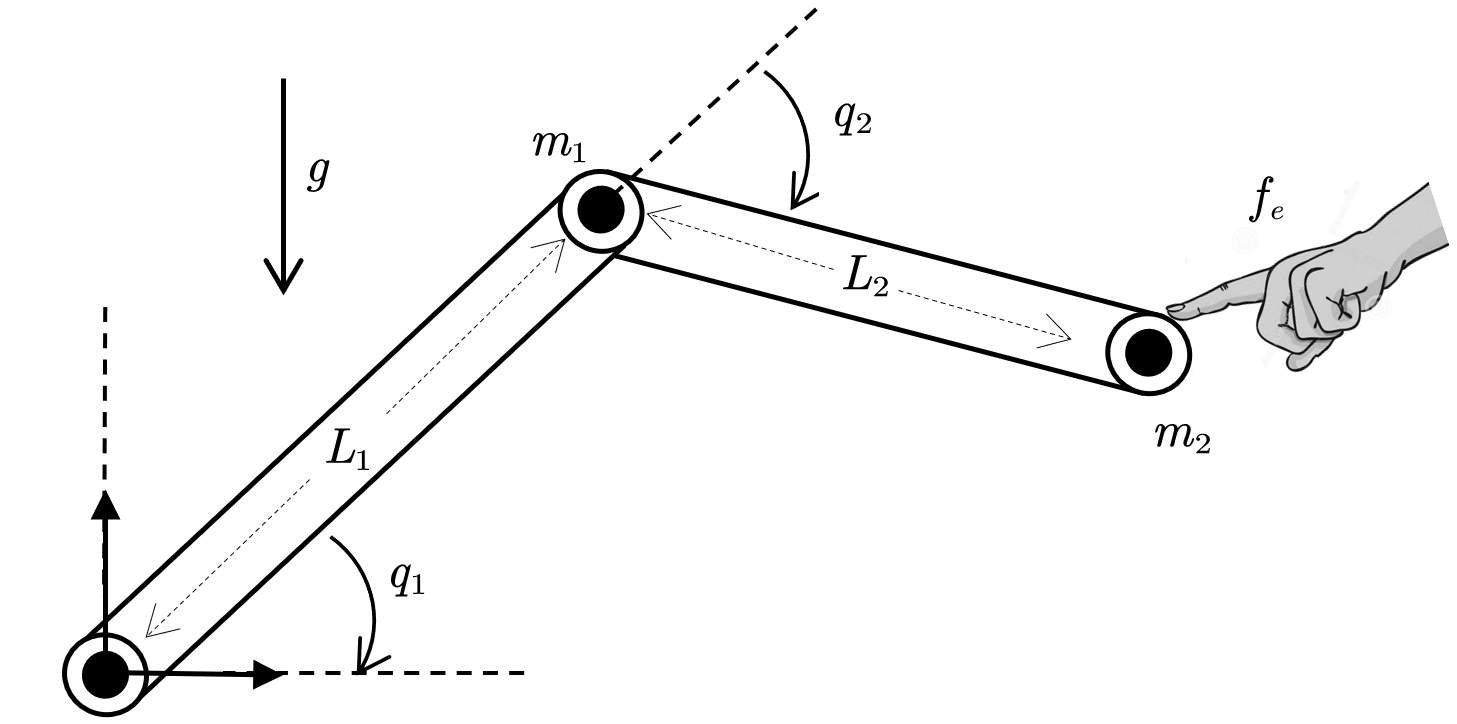}
	\caption{Two-Link robot manipulator.}
	\label{fig_2}
\end{figure}

We consider a robot manipulator whose joint space dynamics equation can be written in the form:
\begin{equation}
	\label{eq1}
	M\left( q \right) \ddot{q}+C\left( q,\dot{q} \right) \dot{q}+G\left( q \right) +F\left( \dot{q} \right) =\tau _c +\tau _e
\end{equation}
where $q$, $\dot{q}$, $\ddot{q}$ are the joint positions, velocities and accelerations of the robot manipulator, respectively. $q=\left[ q_1,q_2,...,q_N \right] ^T$ and $N$ is the number of joints. $M\left( q \right)$ is the $N \times N$ mass matrix, $C\left( q,\dot{q} \right)$ is the $N \times N$ Coriolis and centrifugal forces matrix, $G\left( q \right)$ is the $N \times 1$ gravity vector, and $F\left( \dot{q} \right)$ is the $N \times 1$ vector of frictional forces. $\tau_c$ denotes the control torque acting at the joints, and $\tau_e$ is the external torque from the interaction with the human partner. $F\left( \dot{q} \right)$ and $\tau_e$ together form the time-varying lumped uncertainties of the system. Fig. \ref{fig_2} above shows the structure of a classical two-link robot manipulator interacting physically with a human.

Since it is easier to formulate the trajectory in Cartesian space, we transfer the joint space dynamics of the robot manipulator (\ref{eq1}) into Cartesian space:
\begin{equation}
	\label{eq2}
	M_x\ddot{x}+C_x\dot{x}+G_x+F_x=f_c+f_e
\end{equation}
where $x=\left[ x_1,x_{2},...,x_N \right]$ is the position of the joints in Cartesian space, $M_x$, $C_x$, $G_x$ and $F_x$ are coefficient matrices in Cartesian space, $f_c$ is the control force, and $f_e$ is the external force. The joint space velocities and Cartesian space velocities are related as: 
\begin{equation}
	\label{eq3}
	\dot{x}=J\left( q \right) \dot{q}
\end{equation}
where $J\left( q \right)$ is the Jacobian of the robot manipulator. Inserting (\ref{eq3}) into (\ref{eq1}), the coefficient matrices in Cartesian space can be written as:
$$
M_x=J^{-T}\left( q \right) M\left( q \right) J^{-1}\left( q \right) 
$$
$$
C_x=J ^{-T}\left( q \right)\left( C\left( q,\dot{q} \right) -M\left( q \right) J^{-1}\left( q \right) \dot{J}\left( q \right) \right) J^{-1}\left( q \right) 
$$
$$
G_x=J^{-T}\left( q \right) G\left( q \right) ,\ F_x=J^{-T}\left( q \right) F\left( \dot{q} \right) 
$$
$$
f_c=J^{-T}\left( q \right) \tau _c,\ \ f_e=J^{-T}\left( q \right) \tau _e
$$

Letting $\eta _1=x$ and $\eta _2=\dot{x}$, the Cartesian space model of the robot can be represented as:
\begin{equation}
	\label{eq4}
	\left\{ \begin{array}{l}
		\dot{\eta}_1=\eta _2\\
		\dot{\eta}_2=M_x^{-1}\left( -C_x\dot{x}-G_x-F_x+f_e+f_c \right)\\
	\end{array} \right. 
\end{equation}

For ease of control design, let $\varXi =M_x^{-1}$, $\varGamma \left( x,\dot{x} \right) =M_x^{-1}\left( -C_x-G_x \right)$, $\varPsi \left( x,\dot{x},t \right) =M_x^{-1}\left( -F_x+f_e \right)$, and $u=f_c$, such that the Cartesian space model can be further written as:
\begin{equation}
	\label{eq5}
	\left\{ \begin{array}{l}
		\dot{\eta}_1=\eta _2\\
		\dot{\eta}_2=\varXi u+\varGamma \left( x,\dot{x} \right) +\varPsi \left( x,\dot{x},t \right)\\
	\end{array} \right. 
\end{equation}

\textbf{\emph{Assumption 1}}: There exists a positive constants $\rho$, such that the unknown lumped uncertainties term is bounded by:
\begin{equation}
	\label{eq6}
	\varPsi \left( x,\dot{x},t \right) =M_x^{-1}\left( -F_x+f_e \right)\le \rho 
\end{equation}
where $\rho$ is a positive constant. $\varPsi \left( x,\dot{x},t \right)=[\varPsi_i\left( x,\dot{x},t \right), ..., \varPsi_N \left( x,\dot{x},t \right)]^T$. This assumption means that the lumped uncertainties such as joint errors, friction and external human forces are bounded, which is usually satisfied in real application scenarios.

The control objective of this paper is to develop a fixed-time ISMC scheme to make the joint position $x$ track a certain trajectory $x_r$ which is generated from admittance control. Therefore, the robot manipulator can comply with human forces rather than reject them as external disturbances. Furthermore, we require that the tracking error converges within an arbitrarily small interval of time, regardless of the initial conditions.

\subsection{Preliminaries}
Consider a time-varying differential equation:
\begin{equation}
	\label{eq7}
	\dot{x}=f\left( x,t \right) 
\end{equation}
where $x\in R^n$, and $f:\ R^n\rightarrow R^n$ is a continuous nonlinear function. Assume that the origin of (\ref{eq7}) is the stable equilibrium point and $f\left(0\right)=0$.

\textbf{\emph{Definition 1}}\cite{zuo2015non}: The system (\ref{eq7}) is finite time stable if the origin is Lyapunov stable and any solution $x\left(t\right)$ starting from $x_0$ satisfies $\lim _{t\rightarrow T\left( x_0 \right)}x\left( t \right) \rightarrow 0$ and $x\left( t \right) =0,\forall t >T\left( x_0 \right) 
$, where $T\left( x_0 \right)$ represents the settling time of the system.

\textbf{\emph{Definition 2}}\cite{zuo2015non}: The system (\ref{eq7}) is fixed-time stable if it is globally finite time stable and its settling time $T\left( x_0 \right)$ satisfies $T\left( x_0 \right) <T_{\max}$, where $T_{\max}$ is a positive number.

\textbf{\emph{Lemma 1}} \cite{zuo2016distributed}: The following inequality holds for any real number that satisfies $x_1,x_2,...,x_n>0$ and $0<k<1$:
\begin{equation}
	\label{eq8}
	\sum_{i=1}^n{ x_i ^{k+1}\ge \left( \sum_{i=1}^n{ x_i ^2} \right)}^{\frac{k+1}{2}}
\end{equation}

\textbf{\emph{Lemma 2}} \cite{zuo2016distributed}: The following inequality holds for any real number that satisfies $x_1,x_2,...,x_n>0$ and $k>1$: 
\begin{equation}
	\label{eq9}
	\sum_{i=1}^n{ x_i ^k\ge n^{1-k}\left( \sum_{i=1}^n{ x_i } \right)}^k
\end{equation}

\textbf{\emph{Lemma 3}} \cite{zuo2016distributed}: Consider a scalar differential system below:
\begin{equation}
	\label{eq9_1}
	\dot{y}=\lambda _1\left[ y \right] ^{\alpha}+\lambda _2\left[ y \right] ^{\beta},\ y\left( 0 \right) =y_0
\end{equation}
where $\lambda _1$, $\lambda _2$, $\alpha$, $\beta$ are all positive real number satisfying $\lambda _1, \lambda _2>0$ , $0<\alpha<1$ and $\beta>1$. Then the system (\ref{eq9_1}) is fixed-time stable and the convergence time is independent with respect to the initial state of the system and upper bounded by the design parameters as: 
\begin{equation}
	\label{eq9_2}
	T\left( x_0 \right) <T_{\max}=\frac{1}{\lambda _1\left( 1-\alpha \right)}+\frac{1}{\lambda _2\left(\beta -1 \right)}
\end{equation}

\section{Control design and stability analysis}\label{sec3}
%
%
\subsection{Admittance Trajectory Shaping}
\begin{figure}[h]
	\centering
	\includegraphics[width=2.2in]{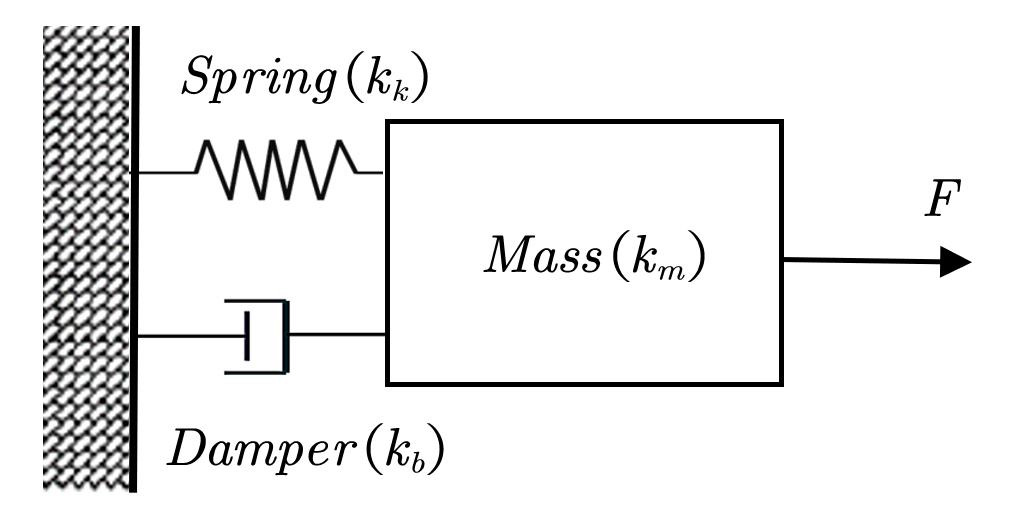}
	\caption{ The mass-spring-damper system.}
	\label{fig_3}
\end{figure}

To comply with external human forces, the system at the contact point can be modeled as a mass-spring-damper system (shown in Fig. \ref{fig_3}). The objective of modeling the system with virtual mass, spring and damper is to make sure the interaction forces are elastic and never vibrate at the contact point. 

Let $\xi _i=x_{r_i}-x_{d_i}, i=1,2,...,N$, where the $x_{r_i}$ is the reference trajectory generated from admittance control and the $x_{d_i}$ is the initial desired trajectory of the robot manipulator. The dynamics for a robot manipulator rendering an impedance can be written as:
\begin{equation}
	\label{eq10}
	k_{m_i}\ddot{\xi}_i+k_{b_i}\dot{\xi}_i+k_{k_i}\xi _i=f_{e_i}
\end{equation}
where $k_{m_i}$, $k_{b_i}$ and $k_{k_i}$ are mass, spring and damping coefficient. $f_e$ is the external human force. By integrating the impedance equation (\ref{eq10}), we obtain $x_{r_i}$ which will be tracked by the controller.

%
%
\subsection{Design of Fixed-time ISMC}

Let the tracking error be written as: 
\begin{equation}
	e=x-x_r
\end{equation}
where $x_r$ is the reference trajectory generated from the admittance control, and $x$ is the real trajectory of robot. The derivative of tracking error $e$ is $\dot{e}=\dot{x}-\dot{x}_r$. To obtain a fixed-time integral sliding mode controller, the error is reconstructed based on Lemma 3: 
\begin{equation}
	\label{eq12}
	s=\dot{e}+k_1\left[ e \right] ^{\alpha}+k_2\left[ e \right] ^{\beta}
\end{equation}
where $k _1$, $k _2$, $\alpha$, $\beta$ are all positive constants satisfying $k _1, k _2>0$ , $0<\alpha<1$ and $\beta>1$. When the sliding surface $s$ converges to zero, (\ref{eq12}) can be rewritten as:
\begin{equation}
	\label{eq12_1}
	\dot{e}=-k_1\left[ e \right] ^{\alpha}-k_2\left[ e \right] ^{\beta}
\end{equation}

According to Lemma 3, the convergence time of (\ref{eq12_1}) is bounded by:
\begin{equation}
	\label{eq12_2}
	T_{s1}\le \frac{1}{k_1\left( 1-\alpha \right)}+\frac{1}{k_2\left( \beta -1 \right)}
\end{equation}

Differentiating (\ref{eq12}) with respect to time and using (\ref{eq5}), we have:
\begin{equation}
	\label{eq13}
	\begin{aligned}
		\dot{s}=\ddot{x}-\ddot{x}_r+k_1\alpha \left[ e \right] ^{\alpha -1}\dot{e}+k_2\beta \left[ e \right] ^{\beta -1}\dot{e}
		=\varXi u+\varGamma\left(x,\dot{x}\right) +\varPsi\left(x,\dot{x},t\right) -\ddot{x}_r+k_1\alpha \left[ e \right] ^{\alpha -1}\dot{e}+k_2\beta \left[ e \right] ^{\beta -1}\dot{e}
	\end{aligned}
\end{equation}

Based on (\ref{eq12}), the integral sliding surface is selected as:
	\begin{equation}
		\label{eq14}
		\begin{aligned}
			\sigma_1 \left( t \right) =s\left( t \right) -s\left( 0 \right) 
			-\int\limits_0^t{\left( \varXi u_0+\varGamma\left(x,\dot{x}\right) -\ddot{x}_r+k_1\alpha \left[ e \right] ^{\alpha -1}\dot{e}+k_2\beta \left[ e \right] ^{\beta -1}\dot{e} \right) dt}
		\end{aligned}
	\end{equation}
where term $s\left(0\right)$ represents the value of the sliding surface when $t=0$, which is added to eliminate the initial error. $u_0$ is the output of the nominal controller, which is designed for the non-disturbance system by another method (e.g., PID, CTC, etc.) and will be presented in the Section 3.4. The derivative of the integral sliding surface can be further written as:
	\begin{equation}
		\label{eq15}
		\begin{aligned}
			\dot{\sigma_1}\left( t \right) =\left( \varXi u+\varGamma\left(x,\dot{x}\right) +\varPsi\left(x,\dot{x},t\right) -\ddot{x}_r+k_1\alpha \left[ e \right] ^{\alpha -1}\dot{e}+k_2\beta \left[ e \right] ^{\beta -1}\dot{e} \right) 
			-\left( \varXi u_0+\varGamma\left(x,\dot{x}\right) -\ddot{x}_r+k_1\alpha \left[ e \right] ^{\alpha -1}\dot{e}+k_2\beta \left[ e \right] ^{\beta -1}\dot{e} \right) 
		\end{aligned}
	\end{equation}

Therefore, we have: 
$$\dot{\sigma_1}\left( t \right)=\varXi \left( u-u_0 \right) +\varPsi \left( x,\dot{x},t \right) $$

Letting $u=u_0+u_{s1}$, where $u_{s1}$ is designed to compensate for disturbances based on $u_0$, we obtain:
\begin{equation}
	\label{eq16}
	\dot{\sigma_1}\left( t \right) =\varXi u_{s1}+\varPsi \left( x,\dot{x},t \right) 
\end{equation}

From (\ref{eq16}), the proposed compensating controller $u_{s1}$ is designed as:
\begin{equation}
	\label{eq17}
	u_{s1}=\varXi ^{-1}\left( -\left( \rho +\varepsilon \right) sign\left( \sigma_1 \right) - k_3\left[ \sigma_1 \right] ^p-k_4\left[ \sigma_1 \right] ^q \right) 
\end{equation}
where the $k_3,k_4>0$, $0<p<1$ and $q>1$. Inserting the compensating controller $u_{s1}$ into  (\ref{eq16}), yields: 
\begin{equation}
	\label{eq18}
	\dot{\sigma_1}\left( t \right) =\varPsi\left( x,\dot{x},t \right)  -\left( \rho +\varepsilon \right) sign\left( \sigma_1 \right) -k_3\left[ \sigma_1 \right] ^p-k_4\left[ \sigma_1 \right] ^q
\end{equation}

and 
\begin{equation}
	\label{eq182}
	\dot{\sigma_{1 _{i}}}\left( t \right) =\varPsi_i\left( x,\dot{x},t \right)  -\left( \rho +\varepsilon \right) sign\left( \sigma_{1 _{i}} \right) -k_3\left[ \sigma_{1 _{i}} \right] ^p-k_4\left[ \sigma_{1 _{i}} \right] ^q
\end{equation}
Consider a Lyapunov function candidate:
\begin{equation}
	\label{eq19}
	V_1=\frac{1}{2}\sum_{i=1}^N{\sigma_{1 _{i}}^{2}}
\end{equation}

Differentiating the Lyapunov function (\ref{eq19}), we have:

	\begin{equation}
		\label{eq20}
		\begin{aligned}
			\dot{V_1}&=\sum_{i=1}^N{\sigma_{1 _{i}}\dot{\sigma_{1 _{i}}}} \\
			&=\sum_{i=1}^N{\sigma_{1 _{i}}\left( \varPsi_i\left( x,\dot{x},t \right)  -\left( \rho +\varepsilon \right) sign\left( \sigma_{1 _{i}} \right) -k_3\left[ \sigma_{1 _{i}} \right] ^p-k_4\left[ \sigma_{1 _{i}} \right] ^q \right)} \\
			&=\sum_{i=1}^N{\left( \sigma _{1 _{i}}\varPsi_i\left( x,\dot{x},t \right)  -\left( \rho +\varepsilon \right) \left| \sigma_{1 _{i}} \right|-k_3\left[ \sigma_{1 _{i}} \right] ^{p+1}-k_4\left[ \sigma_{1 _{i}} \right] ^{q+1} \right)}		
		\end{aligned}
	\end{equation}

By using Assumption 1, we have:
\begin{equation}
	\label{eq20_1}
	\begin{aligned}
		\dot{V_1}\le \sum_{i=1}^N{\left( -k_3\left[ \sigma_{1 _{i}} \right] ^{p+1}-k_4\left[ \sigma_{1 _{i}} \right] ^{q+1} \right)}		
	\end{aligned}
\end{equation}

By using Lemma 1 and Lemma 2, we have: 
\begin{equation}
	\label{eq21}
	\begin{aligned}
		\dot{V_1}&\le -k_3\sum_{i=1}^N{\left| \sigma _{1 _{i}} \right|^{p+1}-}k_4\sum_{i=1}^N{\left| \sigma _{1 _{i}} \right|^{q+1}}\\
		&\le -k_3\sum_{i=1}^N{\left( \left| \sigma _{1 _{i}} \right|^2 \right) ^{\frac{p+1}{2}}}-k_4\sum_{i=1}^N{\left( \left| \sigma _{1 _{i}} \right|^2 \right) ^{\frac{q+1}{2}}}\\
		&\le -k_3\left( \sum_{i=1}^N{\left| \sigma _{1 _{i}} \right|^2} \right) ^{\frac{p+1}{2}}-k_4N^{\frac{1-q}{2}}\left( \sum_{i=1}^N{\left| \sigma _{1 _{i}} \right|^2} \right) ^{\frac{q+1}{2}}\\
		&=-k_32^{\frac{p+1}{2}}\left( V_1 \right) ^{\frac{p+1}{2}}-k_42^{\frac{q+1}{2}}N^{\frac{1-q}{2}}\left( V_1 \right) ^{\frac{q+1}{2}}
	\end{aligned}
\end{equation}

Based on the Lemma 3, the above system is fixed-time stable and the convergence time is upper bounded as:
\begin{equation}
	\label{eq22}
	T_{r1}\le \frac{1}{2^{\frac{p+1}{2}}k_3\left( 1-\frac{p+1}{2} \right)}+\frac{1}{2^{\frac{q+1}{2}}k_4N^{\frac{1-q}{2}}\left( \frac{q+1}{2}-1 \right)}
\end{equation}

\noindent\textbf{\emph{Theorem 1}}. The system (\ref{eq5}) with controller $u_s$ is fixed-time stable and the settling time is bounded by:
\begin{equation}
	\label{eq22_1}
	T\le T_{s1} + T_{r1}+T_{n}
\end{equation}
where $T_{s1}$ and $T_{r1}$ represent the convergence time of the sliding variable $s$ and the system (\ref{eq5}) with $u_{s1}$ as the controller, respectively. $T_n$ represents the convergence time of the system (\ref{eq5}) with $u_0$ as the nominal controller, which will be determined in Section 3.4.

\noindent\textbf{\emph{Proof}}. It is clear that the convergence time of the entire system is smaller than the sum of the convergence times of each component (i.e., the convergence times of the sliding surface, compensating controller and nominal controller). This completes the proof.

\begin{remark}
The value for $\rho$ of the controller in (\ref{eq17}) was selected based on \textbf{Assumption 1}. In practice, we can use experiments to determine the bounded value of the uncertainty $\rho$. Another method which can be applied to approximate the value of parameter of $\rho$ is to use an adaptive technique; for example, we can use an adaptive second-order super-twisting method \cite{van2021robust} to estimate the adaptive gain and eliminate the chattering at the same time.
\end{remark}

\begin{remark}
The use of the `sign' function in the controller (\ref{eq17}) will generate a chattering. To eliminate the chattering, a boundary layer method or second-order super-twisting method can be employed \cite{van2021robust}.
\end{remark}

%
%
\subsection{Nonsingular Problem}

From (\ref{eq13}), when $e=0$ and $\dot{e}\ne 0$ the term $k_1\alpha \left[ e \right] ^{\alpha -1}\dot{e}$ results in a singular problem. To avoid the singularity, the error can be reconstructed as:
\begin{equation}
	\label{eq23}
	s = e+\frac{1}{k_{5}^{m}}\left[ \dot{e}+k_6\left[ e \right] ^n \right] ^{\frac{1}{m}}
\end{equation}
where $k _5$, $k _6$, $m$, $n$ are all positive constants satisfying $k _5, k _6>0$ , $0<m<1$ and $n>1$. When the sliding surface variable converges to zero, (\ref{eq23}) is equal to (\ref{eq12}). Therefore, according to the Lemma 3, the convergence time of (\ref{eq23}) is bounded by:
\begin{equation}
	\label{eq23_1}
	T_{s2}\le \frac{1}{k_5\left( 1-m \right)}+\frac{1}{k_6\left( n-1 \right)}
\end{equation}

Differentiating (\ref{eq23}) with respect to time, we have:
\begin{equation}
	\label{eq24}
	\dot{s}=\dot{e}+\frac{1}{mk_{5}^{m}}\left[ \dot{e}+k_6\left[ e \right] ^n \right] ^{\frac{1}{m}-1}\left( \ddot{e}+k_6n\left[ e \right] ^{n-1} \dot{e}\right) 
\end{equation}

The integral sliding surface is selected as:
	\begin{equation}
		\label{eq25}
		\begin{aligned}
			\sigma_2 \left( t \right) =s\left( t \right) -s\left( 0 \right) 
			-\int_0^t{\left(\dot{e}+\frac{1}{mk_{5}^{m}}\left[ \dot{e}+k_6\left[ e \right] ^n \right] ^{\frac{1}{m}-1}\left( \ddot{e}+k_6n\left[ e \right] ^{n-1} \right) \dot{e} \right)dt}
		\end{aligned}
	\end{equation}

According to the (\ref{eq5}), we have:
\begin{equation}
	\label{eq25_1}
	\ddot{e}=\ddot{x}-\ddot{x_r}=\varXi u+\varGamma \left( x,\dot{x} \right)-\ddot{x_r}
\end{equation}

Inserting (\ref{eq25_1}) into (\ref{eq25}) and letting: 
	\begin{equation}
		\label{eq26}
		\begin{aligned}
			&T_1\left( e,\dot{e} \right) =\frac{1}{mk_{5}^{m}}\left[ \dot{e}+k_6\left[ e \right] ^n \right] ^{\frac{1}{m}-1}\\
			&T_2\left( e \right) = k_6n\left[ e \right] ^{n-1} \dot{e}
		\end{aligned}
	\end{equation}

The derivative of integral sliding surface can be written as:
\begin{equation}
	\begin{aligned}
		\dot{\sigma_2}\left( t \right) =\left(\dot{e}+T_1\left( e,\dot{e} \right)\left( \varXi u+\varGamma\left( x,\dot{x} \right) +\varPsi\left( x,\dot{x},t \right)-\ddot{x}_r+T_2\left( e \right) \right)\right)
		-\left(\dot{e}+T_1\left( e,\dot{e} \right)\left( \varXi u_0+\varGamma\left( x,\dot{x} \right) -\ddot{x}_r+T_2\left( e \right) \right)\right)
	\end{aligned}
\end{equation}

Setting $u=u_0+u_{s2}$, we have:
\begin{equation}
	\label{eq27}
	\dot{\sigma_2}\left( t \right) =T_1\left( e,\dot{e} \right)\left(\varXi u_{s2}+\varPsi \left( x,\dot{x},t \right) \right)
\end{equation}

Again, the controller $u_{s2}$ can be designed as:
\begin{equation}
	u_{s2}=\varXi ^{-1}\left( -\left( \rho +\varepsilon \right) sign\left( \sigma_2 \right) -k_3\left[ \sigma_2 \right] ^p-k_4\left[ \sigma_2 \right] ^q \right) 
\end{equation}

Inserting the compensating controller into  (\ref{eq27}), we have: 
\begin{equation}
	\label{eq28}
	\dot{\sigma_2}\left( t \right) =T_1\left( e,\dot{e} \right)\left(\varPsi\left( x,\dot{x},t \right) -\left( \rho +\varepsilon \right) sign\left( \sigma_2 \right) -k_3\left[ \sigma_2 \right] ^p-k_4\left[ \sigma_2 \right] ^q\right)
\end{equation}

Considering a Lyapunov function candidate:
\begin{equation}
	\label{eq28_1}
	V_2=\frac{1}{2}\sum_{i=1}^N{\sigma_{2 _{i}}^{2}}
\end{equation}

Differentiating the Lyapunov function (\ref{eq19}), we have:
\begin{equation}
	\label{eq28_2}
	\begin{aligned}
		\dot{V_2}&=\sum_{i=1}^N{\sigma_{2 _{i}}\dot{\sigma_{2 _{i}}}} \\
		&=\sum_{i=1}^N{\sigma_{2 _{i}}T_1\left( e,\dot{e} \right)\left( \varPsi_i\left( x,\dot{x},t \right)  -\left( \rho +\varepsilon \right) sign\left( \sigma_{2 _{i}} \right) -k_3\left[ \sigma_{2 _{i}} \right] ^p-k_4\left[ \sigma_{2 _{i}} \right] ^q \right)} \\
		&=T_{1}\left( e,\dot{e} \right)\sum_{i=1}^N{\left( \sigma _{2 _{i}}\varPsi_i\left( x,\dot{x},t \right)  -\left( \rho +\varepsilon \right) \left| \sigma_{2 _{i}} \right|-k_3\left[ \sigma_{2 _{i}} \right] ^{p+1}-k_4\left[ \sigma_{2 _{i}} \right] ^{q+1} \right)}\\
		&\le T_{1}\left( e,\dot{e} \right)\sum_{i=1}^N{\left( -k_3\left[ \sigma_{2 _{i}} \right] ^{p+1}-k_4\left[ \sigma_{2 _{i}} \right] ^{q+1} \right)}\\
		&\le T_{1}\left( e,\dot{e} \right)\left(-k_32^{\frac{p+1}{2}}\left( V_2 \right) ^{\frac{p+1}{2}}-k_42^{\frac{q+1}{2}}N^{\frac{1-q}{2}}\left( V_2 \right) ^{\frac{q+1}{2}}\right)		
	\end{aligned}
\end{equation}

\textbf{\emph{Theorem 2}}. The system (\ref{eq5}) with controller $u_{s2}$ is fixed-time stable and its settling time is bounded by:
\begin{equation}
	\label{eq28_6}
	T\le T_{s2} + T_{r2}+T_{n}+\epsilon \left( \tau \right)
\end{equation}
where $T_{s2}$ and $T_{r2}$ represent the convergence time of sliding variable $s$ and the system (\ref{eq5}) with $u_{s2}$ as the controller, respectively. $T_n$ represent the convergence time of the system (\ref{eq5}) with $u_0$ as the nominal controller.

\textbf{\emph{Proof}}. From (\ref{eq28}), we can see the fixed-time stability depends both on the term $\left(-k_32^{\frac{p+1}{2}}\left( V_2 \right)^{\frac{p+1}{2}}-k_42^{\frac{q+1}{2}}N^{\frac{1-q}{2}}\left( V_2 \right) ^{\frac{q+1}{2}}\right)$ and the term $T_1\left( e,\dot{e} \right)$. According to the work of Van and Ceglarek \cite{van2021robust}, for the case of $\dot{e}+k_6\left[ e \right] ^n\ne 0 $, we divide the state space into two different areas $\varOmega _1=\left\{ \left( e,\dot{e} \right) \left| T\left( e,\dot{e} \right) \ge 1 \right. \right\}$ and $\varOmega _2=\left\{ \left( e,\dot{e} \right) \left| T\left( e,\dot{e} \right) <1 \right. \right\}$. 

i. When the system states are in the area of $\varOmega _1$, we have:
\begin{equation}
	\label{eq28_4}
	\begin{aligned}
		\dot{V_2}\le-k_32^{\frac{p+1}{2}}\left( V_2 \right) ^{\frac{p+1}{2}}-k_42^{\frac{q+1}{2}}N^{\frac{1-q}{2}}\left( V_2 \right) ^{\frac{q+1}{2}}
	\end{aligned}
\end{equation}

$V_2=0$ implies the sliding surface $s=0$. Therefore, according to the Lemma 3, the system states will reach the sliding surface $s=0$ within a fixed time, which is upper bounded as:
\begin{equation}
	\label{eq28_5}
	T_{r2}\le \frac{1}{2^{\frac{p+1}{2}}k_3\left( 1-\frac{p+1}{2} \right)}+\frac{1}{2^{\frac{q+1}{2}}k_4N^{\frac{1-q}{2}}\left( \frac{q+1}{2}-1 \right)}
\end{equation}

ii. When the system states are in the area of $\varOmega _2$, according to (\ref{eq28_2}), the sliding surface $s=0$ is still an attractor. In addition, for the case of $\dot{e}+k_6\left[ e \right] ^n = 0 $, for a given $\tau$, there exists a positive constant $\epsilon \left( \tau \right)$ such that the sliding surface $s_2=0$ can be reached from anywhere in the phase plane within a fixed time $t_r<T_{r2}+\epsilon \left( \tau \right) $. \cite{Li2017OnSC} Therefore, the total setting time $T$ is bounded as (\ref{eq28_6}). This completes the proof.

\begin{remark}
Compared to the design of the integral sliding surfaces of the existing FTISMC \cite{huang2020robust,RN446,Wang_2018}, the proposed controller uses the sliding surface in (\ref{eq25}), which takes the dynamics of the system account in the design of the integral sliding surface. This approach provides two advantages: (1) it allows us to design the nominal controller $u_0$ and the reaching controller $u_s$ separately. This facilitates the design procedure and preserves the advantages of both the nominal controller and ISMC.  
\end{remark}

%
%
\subsection{Design of the Nominal Controller}

The nominal controller is designed to stabilize the system neglecting the lumped uncertainties. To stabilize the whole system within a fixed time, we choose fixed-time backstepping control \cite{Mavriplis2003} as the nominal controller. The dynamics of the robot manipulator without lumped uncertainties can be written as:
\begin{equation}
	\left\{ \begin{array}{l}
		\dot{\eta}_1=\eta _2\\
		\dot{\eta}_2=\varXi u+\varGamma \left( x, \dot{x} \right)\\
	\end{array} \right.  
\end{equation}

Letting $s_1=\eta _1-x_r$, then, $\dot{s}_1=\eta_2-\dot{x_r}$, and we can design the stabilizing function as:
\begin{equation}
	\alpha _s=-\left( \lambda _1s_1+\lambda _2s_{1}^{\alpha}+\lambda _3s_{1}^{\beta} \right) +\dot{x}_r
\end{equation}
where $\lambda _1$, $\lambda _2$, $\lambda _3$, $\alpha$, $\beta$ are all positive constants satisfying $0<\alpha<1$ and $\beta>1$. Thus, $\dot{s}_1$ can be written as:
\begin{equation}
	\dot{s}_1=-\left( \lambda _1s_1+\lambda _2s_{1}^{\alpha}+\lambda _3s_{1}^{\beta} \right)
\end{equation}

Selecting the Lyapunov function candidate as:
\begin{equation}
	V_3=\frac{1}{2}s_{1}^{T}s_1
\end{equation}

The derivative of the Lyapunov function $V_3$ is:

\begin{equation}
	\begin{aligned}
		\dot{V}_3&=s_{1}^{T}\dot{s}_1\\
		&=-s_{1}^{T}\left( \lambda _1s_1+\lambda _2s_{1}^{\alpha}+\lambda _3s_{1}^{\beta} \right) \\
		&=-\lambda _1s_{1}^{T}s_1-\lambda _2\left( s_{1}^{T}s_1 \right) ^{\frac{\alpha +1}{2}}-\lambda _3\left( s_{1}^{T}s_1 \right) ^{\frac{\beta +1}{2}}\\
		&\le -\lambda _2\left( s_{1}^{T}s_1 \right) ^{\frac{\alpha +1}{2}}-\lambda _3\left( s_{1}^{T}s_1 \right) ^{\frac{\beta +1}{2}}\\
		&\le -2^{\frac{\alpha +1}{2}}\lambda _2\left( \frac{1}{2}s_{1}^{T}s_1 \right) ^{\frac{\alpha +1}{2}}-2^{\frac{\beta +1}{2}}\lambda _3\left( \frac{1}{2}s_{1}^{T}s_1 \right) ^{\frac{\beta +1}{2}}\\
		&=-2^{\frac{\alpha +1}{2}}\lambda _2\left( V_3 \right) ^{\frac{\alpha +1}{2}}-2^{\frac{\beta +1}{2}}\lambda _3\left( V_3 \right) ^{\frac{\beta +1}{2}}
	\end{aligned}
\end{equation}

Letting $s_2=\eta _2-\alpha _s $, we have:
\begin{equation}
	\label{eq29_1}
	\dot{s}_2=\dot{\eta}_2-\dot{\alpha}_s=\varXi u_0-\varGamma \left( x,\dot{x} \right) -\dot{\alpha}_s
\end{equation}

Selecting the Lyapunov function candidate as:
\begin{equation}
	V_4=\frac{1}{2}s_{2}^{T}s_2
\end{equation}

To obtain the property of fixed-time convergence, we design the nominal controller as:
\begin{equation}
	\label{eq30}
	u_0=\varXi ^{-1}\left( -\varGamma \left( x,\dot{x} \right) +\dot{\alpha}_s-\lambda _1s_2-\lambda _2s_{2}^{\alpha}-\lambda _3s_{2}^{\beta} \right) 
\end{equation}

Inserting (\ref{eq30}) into (\ref{eq29_1}), we have:
\begin{equation}
	\dot{s}_2=-\lambda _1s_2-\lambda _2s_{2}^{\alpha}-\lambda _3s_{2}^{\beta}
\end{equation}

The derivative of the Lyapunov function $V_2$ is:
\begin{equation}
	\begin{aligned}
		\dot{V}_4&=s_{2}^{T}\dot{s}_2\\
		&=s_{2}^{T}\left( -\lambda _1s_2-\lambda _2s_{2}^{\alpha}-\lambda _3s_{2}^{\beta} \right) \\
		&\le -2^{\frac{\alpha +1}{2}}\lambda _2\left( V_4 \right) ^{\frac{\alpha +1}{2}}-2^{\frac{\beta +1}{2}}\lambda _3\left( V_4 \right) ^{\frac{\beta +1}{2}}
	\end{aligned}
\end{equation}

Choosing the Lyapunov function candidate of system as:
\begin{equation}
	V_n=V_3+V_4
\end{equation}

The derivative of Lyapunov function $V$ is:
\begin{equation}
	\begin{aligned}
		\dot{V_n}&\le-2^{\frac{\alpha +1}{2}}\lambda _2\left( V_3 \right) ^{\frac{\alpha +1}{2}}-2^{\frac{\beta +1}{2}}\lambda _3\left( V_3 \right) ^{\frac{\beta +1}{2}}-2^{\frac{\alpha +1}{2}}\lambda _2\left( V_4 \right) ^{\frac{\alpha +1}{2}}-2^{\frac{\beta +1}{2}}\lambda _3\left( V_4 \right) ^{\frac{\beta +1}{2}}\\
		&\le -2^{\frac{\alpha +1}{2}}\lambda _2\left( V_n \right) ^{\frac{\alpha +1}{2}}-2^{\frac{\beta +1}{2}}\lambda _3\left( V_n \right) ^{\frac{\beta +1}{2}}
	\end{aligned}
\end{equation}

According to the Lemma 3, when applying the proposed nominal controller $u_0$, the system without lumped uncertainties is fixed-time stable and the convergence time is bounded as:
\begin{equation}
	T_n\le \frac{2}{\lambda _22^{\frac{\alpha +1}{2}}\left( 1-\alpha \right)}+\frac{2}{\lambda _32^{\frac{\beta +1}{2}}\left( \beta -1 \right)}
\end{equation}

\begin{remark}
It is noted that the nominal controller can be designed using other controllers, e.g, PID, CTC and Backstepping. However, in this paper, we use the fixed-time backstepping controller to achieve global fixed-time convergence for the system.
\end{remark}
%
%

\section{Simulation and results}\label{sec4}

In the simulation, we consider a two-link robot in the horizontal plane \cite{craig1987adaptive}. The dynamics of the robot are described as:
	\begin{equation}
		\label{eq31}
		\begin{aligned}
			\tau _1&=m_2l_{2}^{2}\left( \ddot{q}_1+\ddot{q}_2 \right) +m_2l_1l_2c_2\left( 2\ddot{q}_1+\ddot{q}_2 \right) +\left( m_1+m_2 \right) l_{1}^{2}\ddot{q}_1
			-m_2l_1l_2s_2\dot{q}_{2}^{2}-2m_2l_1l_2s_2\dot{q}_1\dot{q}_2+m_2l_2gc_{12}+\left( m_1+m_2 \right) l_1gc_1\\
			\tau _2&=m_2l_1l_2c_2\ddot{q}_1+m_2l_1l_2s_2\dot{q}_{1}^{2}+m_2l_2gc_{12}+m_2l_{2}^{2}\left( \ddot{q}_1+\ddot{q}_2 \right) 
		\end{aligned}
	\end{equation}
where $c_i=\cos \left( q_i \right) $, $c_{ij}=\cos \left( q_i+q_j \right) $, $s_i=\sin \left( q_i \right) $, and $s_{ij}=\sin \left( q_i+q_j \right)$, $i,j=1,2$. The value of masses are $m_1=1.5kg$, and $m_2=1.0kg$. The length of the links are $l_1=l_2=0.3m$. Based on the (\ref{eq1}), the mass matrix $M\left( q \right)$, Coriolis and centrifugal forces matrix $C\left( q,\dot{q} \right)$ and the gravity matrix $G\left( q \right)$ are given as:
	\begin{equation}
		\label{eq32}
		M\left( q \right) =\left[ \begin{matrix}
			m_2l_{2}^{2}+2m_2l_1l_2c_2+\left( m_1+m_2 \right) l_{1}^{2}&		m_2l_{2}^{2}+m_2l_1l_2c_2\\
			m_2l_{2}^{2}+m_2l_1l_2c_2&		m_2l_{2}^{2}\\
		\end{matrix} \right] 
	\end{equation}
\begin{equation}
	\label{eq33}
	C\left( q,\dot{q} \right) =\left[ \begin{matrix}
		-2m_2l_1l_2s_2\dot{q}_2&		-m_2l_1l_2s_2\\
		m_2l_1l_2s_2\dot{q}_1&		0\\
	\end{matrix} \right] 
\end{equation}
\begin{equation}
	\label{eq34}
	G\left( q \right) =\left[ \begin{array}{c}
		m_2l_2gc_{12}+\left( m_1+m_2 \right) l_1gc_1\\
		m_2l_2gc_{12}\\
	\end{array} \right] 
\end{equation}

The bounded disturbance term is given as:
\begin{equation}
	\label{eq35}
	F\left( q \right) =\left[ \begin{array}{c}
		2c_1s_2+5c_{1}^{2}\\
		-2c_1s_2-5c_{1}^{2}\\
	\end{array} \right] 
\end{equation}

The Jacobian of the robot manipulator is given as:
	\begin{equation}
		\label{eq35_1}
		J\left( q \right) =\left[ \begin{matrix}
			-l_1s_{1} -l_2s_{12}&		-l_2s_{12}\\
			l_1c_1 +l_2c_{12}&		l_2c_{12}\\
		\end{matrix} \right] 
	\end{equation}

Assume the desired trajectory \cite{tee2010adaptive} of the two joints is:
\begin{equation}
	\label{eq11}
	\begin{aligned}
		x_{d1}\left( t \right) &= 0.14\cos \left( 0.5t \right) \\
		x_{d2}\left( t \right) &=0.14\sin \left( 0.5t \right) 	
	\end{aligned}
\end{equation}
\begin{figure}[b]
	\centering
	\begin{minipage}[t]{0.45\textwidth}
		\centering
		\includegraphics[width=0.72\textwidth]{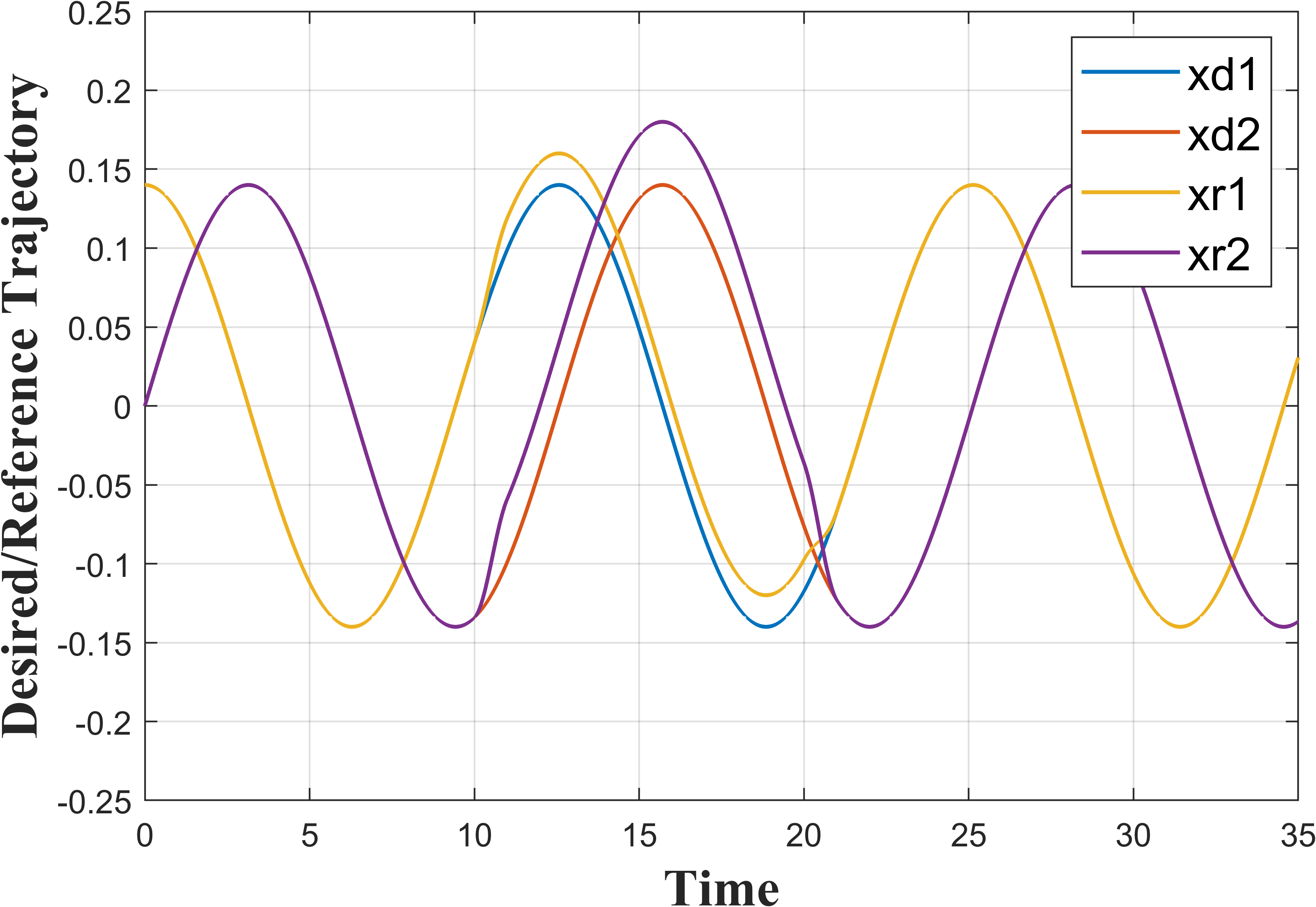}
		\caption{Desired/reference trajectory.}
		\label{fig_4}
	\end{minipage}
	\begin{minipage}[t]{0.45\textwidth}
		\centering
		\includegraphics[width=0.68\textwidth]{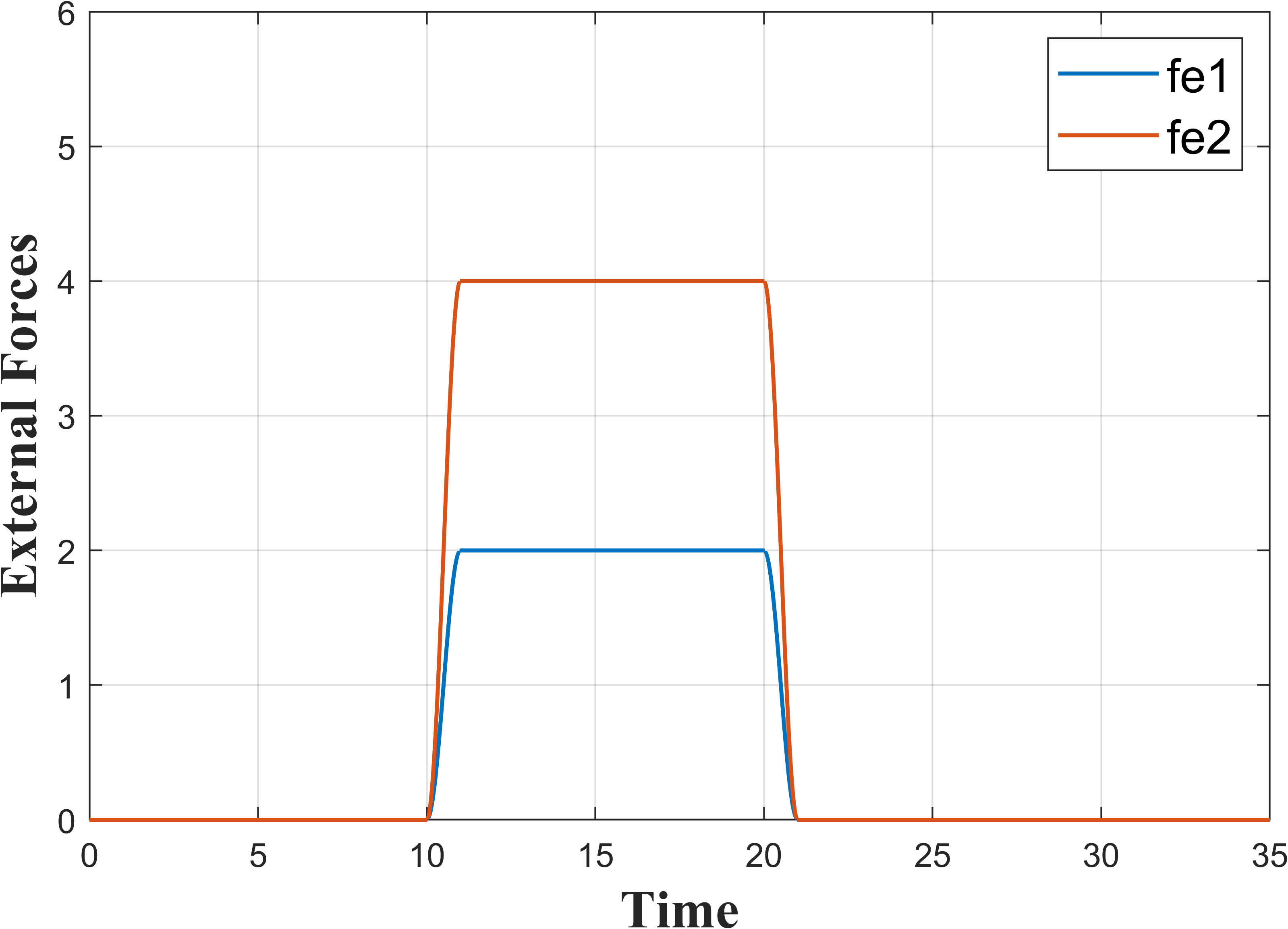}
		\caption{External human forces.}
		\label{fig_5}
	\end{minipage}
\end{figure}

The external human forces, as shown in Fig. \ref{fig_5}, are given by \cite{tee2010adaptive}:
	\begin{equation}
		\label{eq36}
		fe_i\left( t \right) \left\{ \begin{array}{l}
			0\ \ \ \ \ \ \ \ \ \ \ \ \ \ \ \ \ \ \ \ \ \ \ \ \ \ \ \ \ \ \ t<10\ or\ t\ge 21\\
			a_i\left( 1-\cos \pi t \right) \ \ \ \ \ \ \ \ \ 10\le t<11\\
			2a_i\ \ \ \ \ \ \ \ \ \ \ \ \ \ \ \ \ \ \ \ \ \ \ \ \ \ \ 11\le t<20\\
			a_i\left( 1+\cos \pi t \right) \ \ \ \ \ \ \ \ \ 20\le t<21\\
		\end{array} \right. 
	\end{equation}

The external human forces are applied when $t=10s$ and removed at $t=21s$. By applying admittance control, the reference trajectory of the two joints $x_{r1}$ and $x_{r2}$ can be derived by integrating equation (\ref{eq10}) twice. Fig. \ref{fig_4} shows both the desired trajectory and the reference trajectory. We can see the two trajectories are the same before the external forces are applied. When $t>10s$, the reference trajectory, which complies with thr external human forces, is different from the desired trajectory. When $t>21s$, the external human forces are removed from the contact point and the two trajectories coincide again.
\begin{figure}
	\centering
	\includegraphics[width=7in]{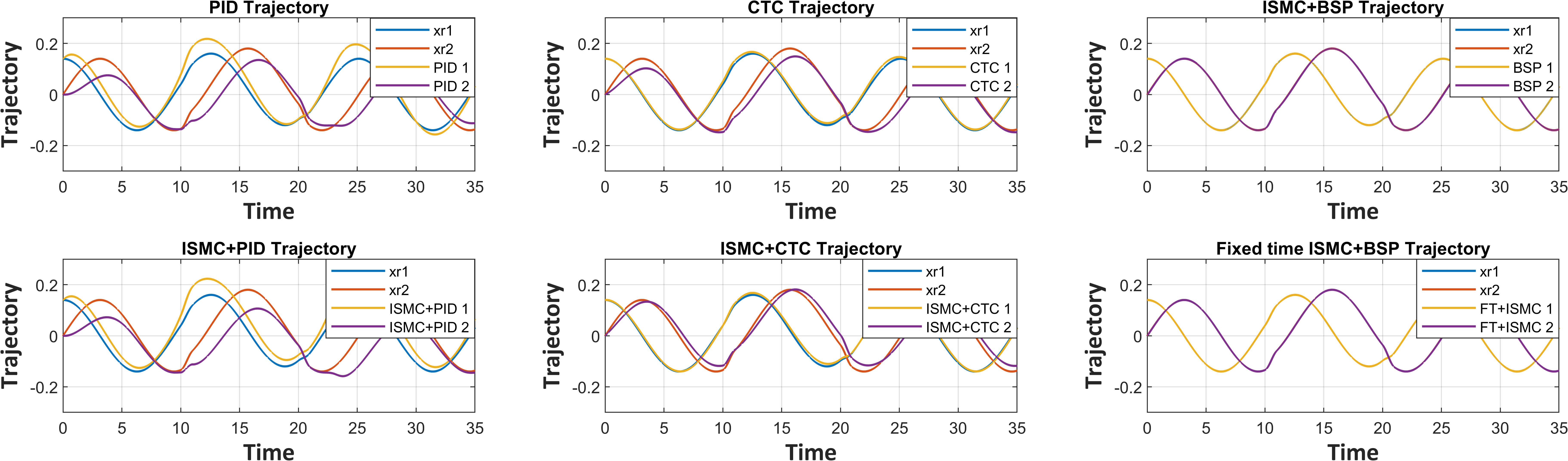}
	\caption{Trajectories obtained with the different controllers.}
	\label{fig_6}
\end{figure}
\begin{figure}[h]
	\centering
	\begin{minipage}[t]{0.495\textwidth}
		\centering
		\includegraphics[width=1\textwidth]{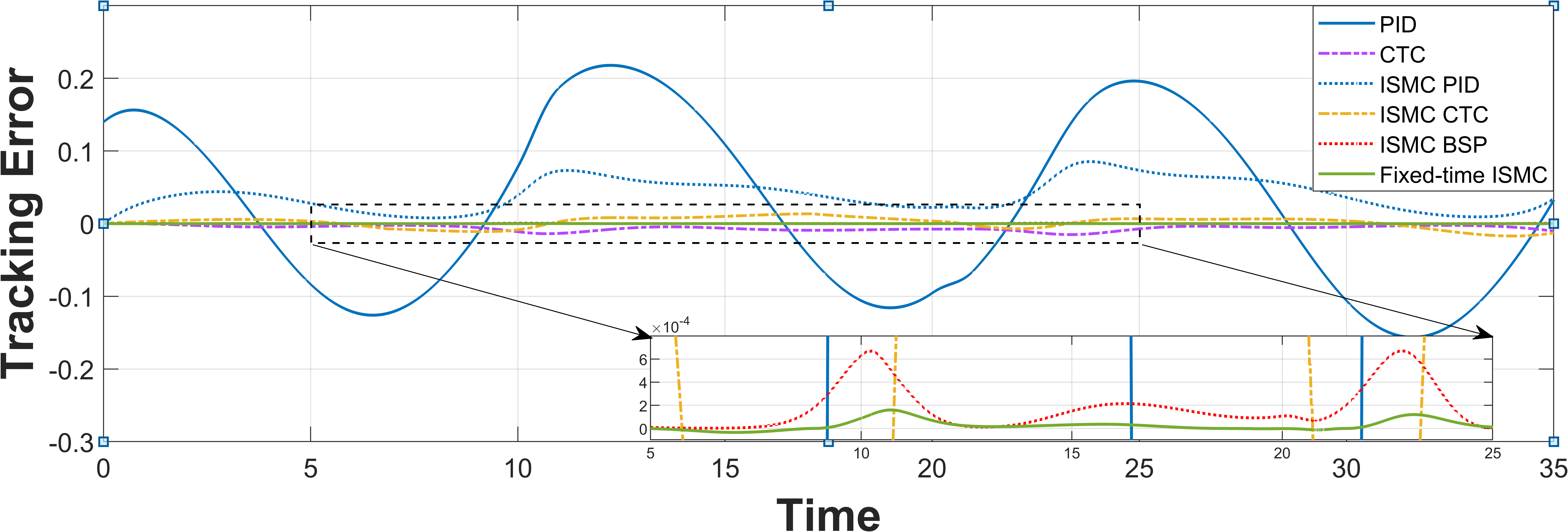}
		\caption{Tracking error for joint 1.}
		\label{fig_7}
	\end{minipage}
	\begin{minipage}[t]{0.495\textwidth}
		\centering
		\includegraphics[width=1\textwidth]{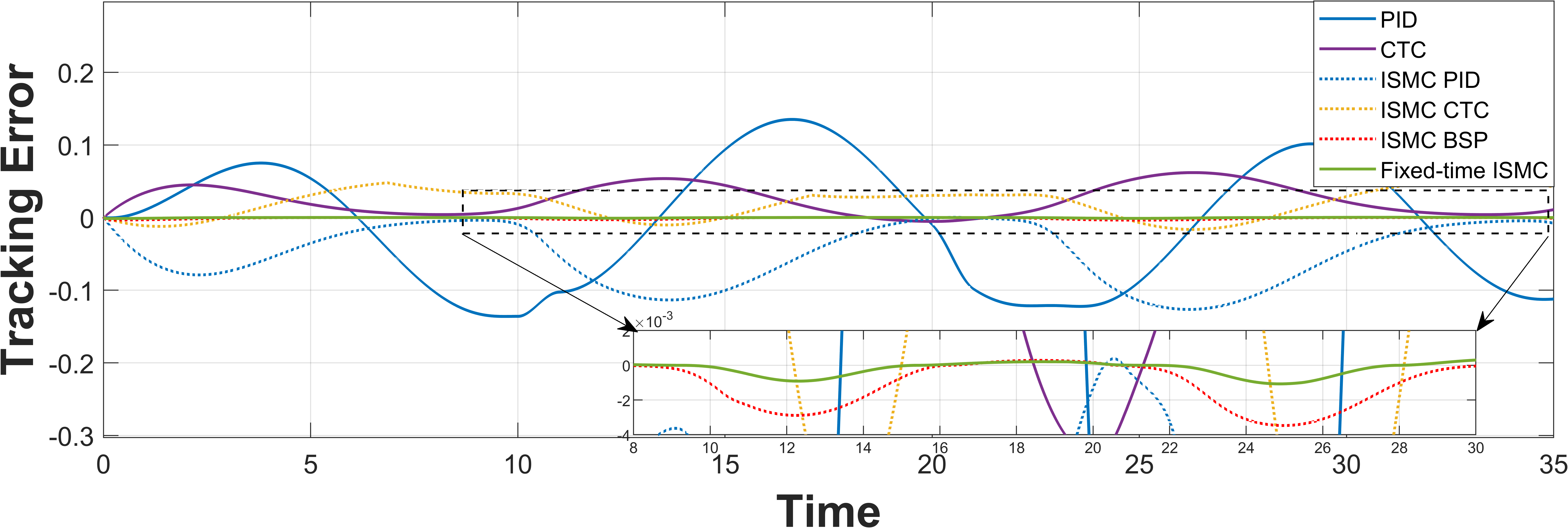}
		\caption{Tracking error for joint 2.}
		\label{fig_8}
	\end{minipage}
\end{figure}

\begin{figure}[h]
	\centering
	\begin{minipage}[t]{0.495\textwidth}
		\centering
		\includegraphics[width=1\textwidth]{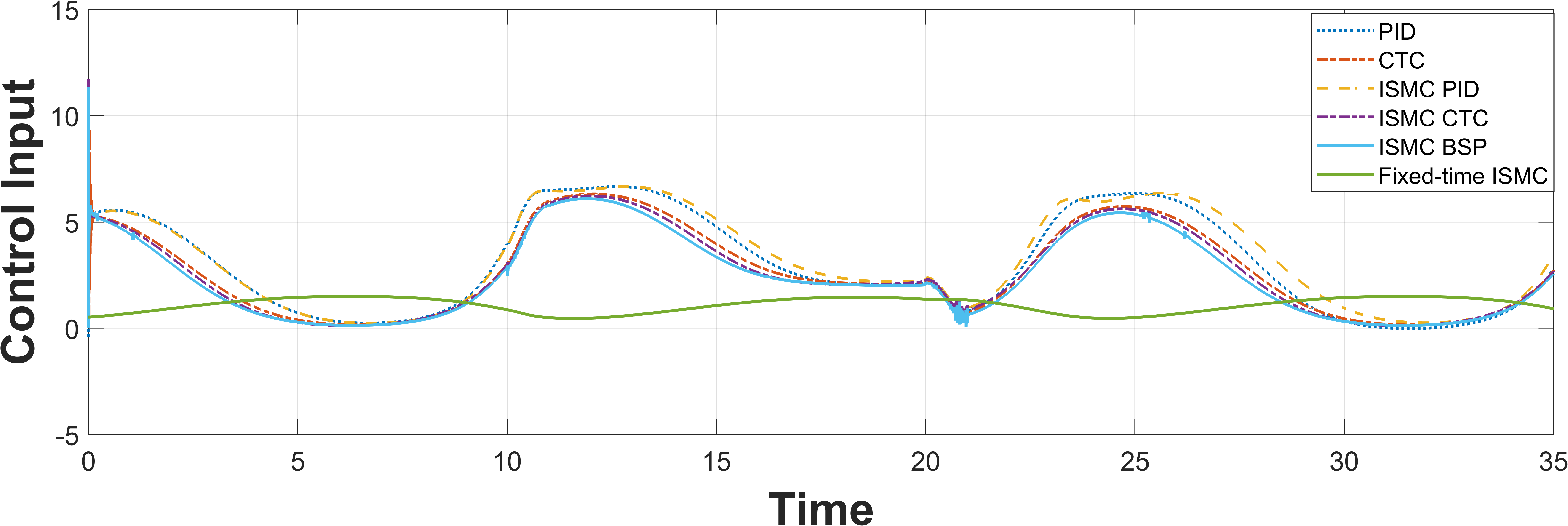}
		\caption{Control input for joint 1.}
		\label{fig_9}
	\end{minipage}
	\begin{minipage}[t]{0.495\textwidth}
		\centering
		\includegraphics[width=1\textwidth]{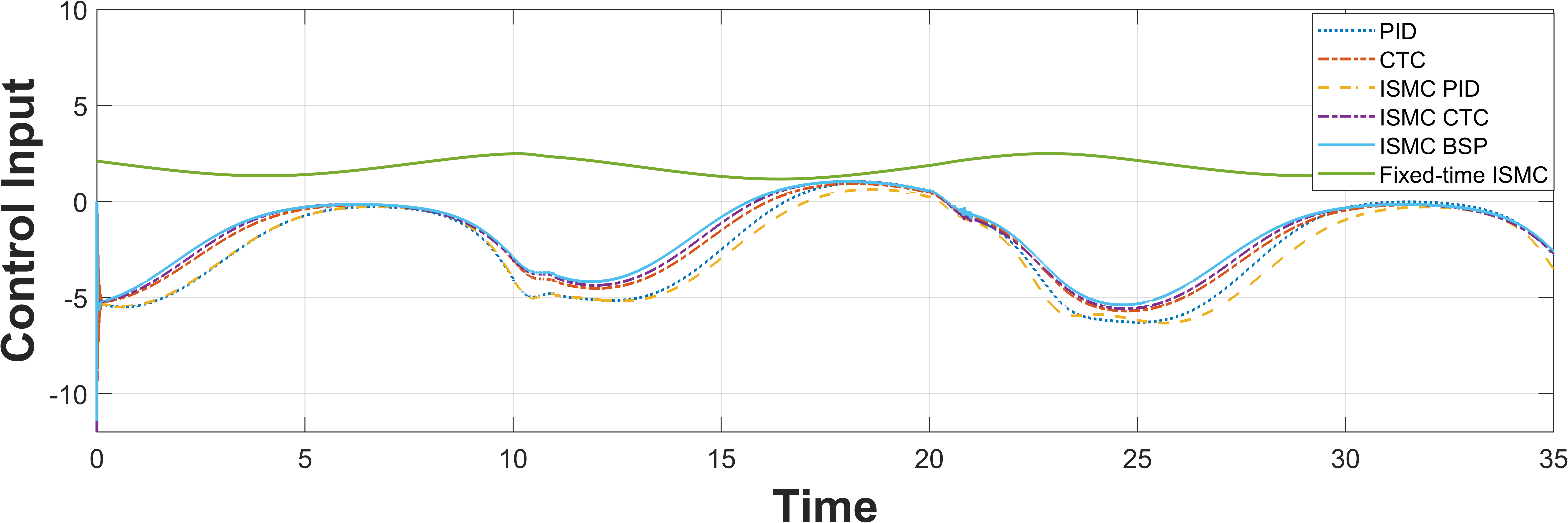}
		\caption{Control input for joint 2.}
		\label{fig_10}
	\end{minipage}
\end{figure}

To verify the effectiveness of the proposed controller, we compare it with PID, computed torque control (CTC), traditional ISMC with PID as the nominal controller (ISMC PID), traditional ISMC with CTC as the nominal controller (ISMC CTC), and traditional ISMC with BSP as the nominal controller (ISMC BSP). Assume the initial position of the end-effector is $q\left(0\right)=\left[0.5236,2.0944\right]^T$, and the initial value of the task variable is  $x\left(0\right)=\left[0,0\right]^T$. The design parameters of the admittance control are $k_{m_i}=20$, $k_{b_i}=20$ and $k_{k_i}=100$, $i=1,2$. The PID parameters are $k_p=300$, $k_d=400$, $k_i=10$. The parameters of BSP are $\lambda_1=3$, $\lambda_2=20$, $\lambda_3=50$, $\alpha=\frac{5}{7}$, $\beta=\frac{5}{3}$. These parameters are selected based on a trial-and-error procedure and by experience. The design parameters of the proposed controller are $k_1=k_3=20$, $k_2=k_4=50$, $m=p=\frac{5}{7}$, $n=q=\frac{5}{3}$. 

The trajectories of the benchmark controllers and the proposed fixed-time ISMC controller with BSP as the nominal controller are shown in Fig. \ref{fig_6}, and the corresponding Root Mean Square Error (RMSE) of tracking errors are reported in Table 1. We can see that the PID controller has the poorest tracking performance. The performances of CTC is a little better than PID. The performance of ISMC PID and ISMC CTC are better than pure PID and CTC, respectively, which means the tracking performance of the controller is improved by adding the integral sliding surface. The ISMC BSP and proposed fixed-time ISMC controller have the best tracking performance. They almost achieve perfectly tracking of the reference trajectory, which makes it difficult to compare them in Fig. \ref{fig_6}. However, as can be seen in Fig. \ref{fig_7} and Fig. \ref{fig_8}, it is clear that the proposed controller has both the smallest tracking error and shortest convergence time. Note that the proposed controller is global fixed-time convergent because both the integral sliding surface, reaching controller and the nominal controller can converge within a fixed time. In addition, from Fig. \ref{fig_9} and Fig. \ref{fig_10}, which show the control inputs of the two joints, it can be seen that the proposed solution provides the smoothest and most efficient control input.

\begin{center}
\label{table1}
	\begin{table}[h]
		\caption{The RMSE performance of each controller.}
		\setlength{\tabcolsep}{5.3mm}{
			\begin{tabular}{ccccccc}
				\toprule
				\makecell[c]{Joint} &\makecell[c]{PID}& \makecell[c] {CTC}  &\makecell[c]{ISMC+PID} &\makecell[c]{ISMC+CTC} &\makecell[c]{ISMC+BSP} &\makecell[c]{FTISMC+BSP}\\				
				\midrule
				\multirowcell{0.1}{1}&0.1211&0.0070&0.0446&0.0064&$2.0901\times 10^{-4}$&$5.7841\times 10^{-5}$\\
				\midrule
				\multirowcell{0.1}{2}&0.0845&0.0314&0.0644&0.0220&0.0016&$4.4889\times 10^{-4}
				$\\			
				\bottomrule
		\end{tabular}}
	\end{table}
\end{center}

\section{Conclusions}\label{sec5}

In this paper, a fixed-time ISMC controller based on admittance control has been proposed for robot manipulators such that the robot can comply with the external human forces rather than reject them as disturbances. Furthermore, the convergence time of the system can be predefined regardless of the initial conditions. The simulation results show that the proposed controller can track the reference trajectory precisely with faster convergence of the tracking error. In future work, a fixed-time force observer of external human forces will be discussed and integrated into the system. Safety constraints of the system will also be considered. The whole system will be tested on a robotics platform such as the Baxter robot to further validate the controller's effectiveness and evaluate real world performance.






\bibliography{wileyNJD-AMA}%

\end{document}